\documentclass[PCJ,Unicode,screen]{cedram}

\usepackage[natbib=true,mincitenames=1,maxcitenames=2,maxbibnames=20,minbibnames=20,uniquelist=false,uniquename=false,style=apa]{biblatex}
\usepackage{lipsum}
\usepackage{setspace}
\usepackage{amsmath, amssymb}
\usepackage{graphicx}
\usepackage{float}
\usepackage{subcaption}
\usepackage{booktabs}
\usepackage{tabularx}
\usepackage{microtype}

\providecommand*{\noopsort}[1]{}

\DOI{10.5802/fake.doi} % this is the DOI of the preprint on the preprint server
\PCIcorresp{david.roqui@ensea.fr}
\title[Multimodal Approach to Heritage Preservation]{A Multimodal Approach to Heritage Preservation in the Context of Climate Change}

\author{\firstname{David} \lastname{ROQUI}}
\address{ETIS laboratory, Cergy university}
\address{Fondation des sciences du patrimoine (FSP)}
\email[D. ROQUI]{david.roqui@ensea.fr}

\author{\firstname{Adèle} \lastname{CORMIER}}
\address{Research and Restoration Center for the Museums of France (C2RMF)}
\address{Epitopos}
\email[A. CORMIER]{adele.cormier@etu.chimieparistech.psl.eu}

\author{\firstname{Nistor} \lastname{GROZAVU}}
\address{Equipes traitement de l'information et systèmes (ETIS)}
\email[N. Grozavu]{nistor.grozavu@cyu.fr}

\author{\firstname{Ann} \lastname{BOURGES}}
\address{Research and Restoration Center for the Museums of France (C2RMF)}
\email[A. Bourges]{ann.bourges@culture.gouv.fr}


\begin{abstract}
\noindent Cultural heritage sites face accelerating degradation due to climate change, yet traditional monitoring relies on unimodal analysis (visual inspection or environmental sensors alone) that fails to capture the complex interplay between environmental stressors and material deterioration. We propose a lightweight multimodal architecture that fuses sensor data (temperature, humidity) with visual imagery to predict degradation severity at heritage sites.

\noindent Our approach adapts PerceiverIO with two key innovations: (1) simplified encoders (64D latent space) that prevent overfitting on small datasets (n=37 training samples), and (2) Adaptive Barlow Twins loss that encourages modality complementarity rather than redundancy. On data from Strasbourg Cathedral, our model achieves 76.9\% accuracy, a 43\% improvement over standard multimodal architectures (VisualBERT, Transformer) and 25\% over vanilla PerceiverIO.

\noindent Ablation studies reveal that sensor-only achieves 61.5\% while image-only reaches 46.2\%, confirming successful multimodal synergy. A systematic hyperparameter study identifies an optimal moderate correlation target ($\tau$=0.3) that balances alignment and complementarity, achieving 69.2\% accuracy compared to other $\tau$ values ($\tau$=0.1/0.5/0.7: 53.8\%, $\tau$=0.9: 61.5\%). This work demonstrates that architectural simplicity combined with contrastive regularization enables effective multimodal learning in data-scarce heritage monitoring contexts, providing a foundation for AI-driven conservation decision support systems.
\end{abstract}
\usepackage{etoolbox}
\makeatletter
\AtBeginDocument{%
  \@ifpackageloaded{lineno}{\nolinenumbers}{}%
}
\makeatother

\begin{document}

\maketitle
\onehalfspacing

\section*{Introduction}

\noindent Climate change accelerates the degradation of cultural heritage worldwide through temperature fluctuations, humidity cycles, and extreme weather events. Traditional conservation relies on periodic expert inspections, a reactive approach insufficient for the pace of climate-driven deterioration. Automated monitoring systems could enable proactive interventions, yet heritage preservation presents unique machine learning challenges. Datasets rarely exceed one hundred samples due to limited site access, expensive expert annotation, and slow degradation timescales spanning years.

\noindent Degradation results from complex interactions between environmental stressors and material properties that visual inspection alone cannot capture. Temperature variations cause thermal expansion weakening structural bonds, while humidity drives salt crystallization within porous materials. This motivates multimodal approaches fusing sensor data with weathering monitoring through imagery. However, existing architectures like VisualBERT \cite{li_visualbert_2019}, UNITER \cite{uniter}, and FLAVA \cite{singh_flava_2022} achieve strong performance on large-scale vision-language benchmarks but fail on specialized small-scale tasks. Pre-trained representations from general images do not transfer to scientific heritage imaging, while high parameter counts cause severe overfitting on limited training sets.

\noindent We propose a lightweight multimodal architecture adapted for data-scarce heritage monitoring. Our approach modifies PerceiverIO \cite{jaegle_perceiver_2022} through two key innovations. First, we replace complex encoders with simple linear projections, reducing parameters to match dataset size and prevent memorization. Second, we introduce Adaptive Barlow Twins loss that encourages modality complementarity rather than redundancy. Unlike standard fusion methods promoting identical representations, our partial correlation target preserves modality-specific information while maintaining semantic coherence.

\noindent We validate this approach on Strasbourg Cathedral monitoring data combining environmental sensors with surface imagery across five degradation classes. Through systematic ablation studies and hyperparameter analysis, we investigate how architectural simplification affects generalization, how modalities contribute individually versus combined, and what balance between alignment and complementarity optimizes performance.

\noindent The paper proceeds as follows. Section 2 reviews related work. Section 3 details our architecture and loss formulation. Section 4 describes the dataset and evaluation protocol. Sections 5 and 6 present results and discussion. Section 7 concludes with future directions.

\section*{Related Work}

\subsection*{Multimodal Learning Architectures}

\noindent Early multimodal approaches relied on modality-specific feature extractors followed by concatenation \cite{ngiam2011multimodal}. Convolutional neural networks for images \cite{cnn} and recurrent networks for sequences \cite{hochreiter1997long} processed each modality independently before late fusion through fully connected layers. However, this strategy fails to capture cross-modal interactions during feature learning, limiting representational power.

\noindent The transformer architecture \cite{attention} revolutionized multimodal learning by enabling attention-based fusion. VisualBERT \cite{li_visualbert_2019} extended BERT \cite{Bert} to vision-language tasks through co-attentional layers, achieving strong performance on visual question answering and image captioning. UNITER \cite{uniter} and FLAVA \cite{singh_flava_2022} further improved cross-modal alignment through contrastive pre-training on large-scale image-text pairs. Vision Transformers \cite{vit} demonstrated that pure attention mechanisms could match or exceed convolutional architectures on image classification when sufficient training data is available.

\noindent Despite their success on large-scale benchmarks, these models face critical limitations in specialized domains. First, pre-training on general vision-language corpora does not transfer effectively to scientific imaging modalities like multispectral sensors or microscopy. Second, model complexity requires datasets with tens of thousands of samples to avoid overfitting, far exceeding typical heritage monitoring budgets. Third, these architectures assume semantic alignment between modalities, whereas our task requires preserving complementarity between environmental sensors and visual evidence.

\subsection*{Fusion Strategies for Heterogeneous Modalities}

\noindent The choice of fusion strategy critically impacts multimodal performance. Early fusion concatenates raw inputs before processing \cite{mdl}, enabling joint feature learning but increasing dimensionality and computational cost. Late fusion combines predictions from modality-specific models \cite{karpathy_large-scale_2014}, preserving specialization but missing cross-modal interactions during training. Intermediate fusion balances these trade-offs through hierarchical integration at multiple network depths \cite{poria2015deep}.

\noindent Perceiver \cite{perceiver} introduced a paradigm shift by mapping diverse input modalities to a shared latent space through iterative cross-attention. This approach handles variable-sized inputs and scales linearly with input length rather than quadratically like standard transformers. PerceiverIO \cite{jaegle_perceiver_2022} extended this framework with flexible output decoders, enabling task-specific predictions while maintaining architectural generality. However, the original Perceiver design targets large-scale pre-training scenarios and requires adaptation for small-data regimes.

\subsection*{Self-Supervised Learning for Multimodal Representations}

\noindent Recent work explores contrastive objectives for learning aligned multimodal representations without explicit labels. CLIP \cite{radford2021learning} trains vision and language encoders to maximize similarity between corresponding image-text pairs while minimizing similarity for mismatched pairs. Barlow Twins \cite{zbontar2021barlow} reduces redundancy between augmented views by decorrelating their representations, avoiding collapse without requiring negative samples.

\noindent These methods assume modalities provide redundant views of the same semantic content. Heritage monitoring violates this assumption: sensors capture environmental causes while images reveal material effects. Our work adapts Barlow Twins from view-invariance to modality-complementarity, encouraging decorrelation rather than alignment.

\subsection*{Machine Learning for Heritage Preservation}

\noindent AI applications in cultural heritage have primarily focused on digital reconstruction and damage detection. Generative adversarial networks restore degraded artworks \cite{elgammal_can_2017}, while convolutional networks detect cracks in historical structures from visual inspection \cite{dais2021automatic}. However, these approaches operate on single modalities and ignore environmental context.

\noindent Recent work explores multimodal heritage monitoring by combining visual surveys with climate data. Cabral et al. \cite{cabral2020automated} fuse thermal imaging with structural sensors for building assessment but rely on large annotated datasets unavailable for most sites. \noindent Grilli and Remondino \cite{grilli20193d} integrate photogrammetry with environmental logging yet analyze modalities separately rather than jointly. To our knowledge, no prior work addresses the joint challenges of multimodal fusion and extreme data scarcity in heritage contexts.

\subsection*{Positioning of Our Work}

\noindent This work is at the intersection of three research areas: small-data deep learning, contrastive multimodal fusion, and heritage science. We adapt PerceiverIO for data-scarce scenarios through architectural simplification, drawing inspiration from network pruning \cite{han2015learning} and knowledge distillation \cite{hinton2015distilling} that demonstrate smaller models can match or exceed larger ones when training data is limited. Our Adaptive Barlow Twins loss extends contrastive learning from view-invariance to modality-complementarity, addressing the gap between existing self-supervised methods and heterogeneous sensor-image fusion. Finally, we provide a benchmark of state-of-the-art multimodal architectures on heritage monitoring data, establishing baselines for future research in this domain.

\section*{Dataset}

\noindent Collected data comes from three French heritage sites (Bibracte archaeological site, Strasbourg Cathedral, and the Saint-Pierre Chapel). These data are currently divided into two modalities : continuous text data from sensors and images ponctuously collected on sites. Several data collection campaigns have already been conducted and occur every six months to monitor weathering evolution. Images are transformed into weathering maps with different layers. It is important to note that for this first article, data used are the first campaign (T0) data and the T0 data from Strasbourg Cathedral. However, other campaigns will take place, allowing for the creation of T1, T2, etc., improving the model's efficiency. Collected data from climatic sensors have good quality, with data collected regularly and without missing data or outlier values. Figures 2 to 4 show examples of what the sensor and image modality data may look like.

\begin{figure}[ht]
\centering
\includegraphics[width=1\linewidth]{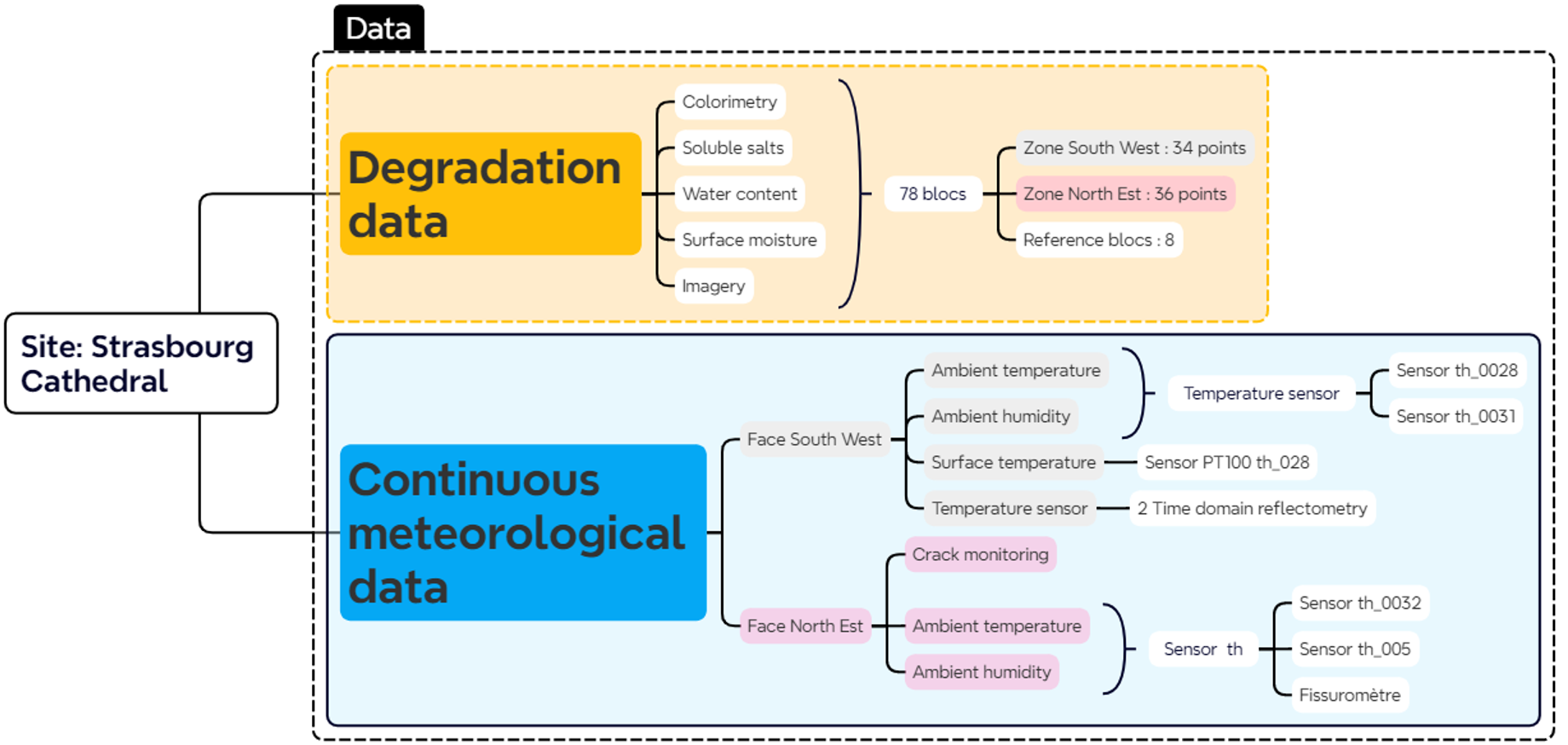}
\caption{Schema of the dataset}
\label{fig:data_fig}
\end{figure}

\subsection*{Climatic continuous data}

\noindent This dataset from climatic and crack sensors represents the text modality. On the three sites, 16 thermohygrometer-sensors continuously record three parameters every hour : temperature (°C), relative humidity (\%) and surface temperature (°C). The data is sent directly to a platform connected to the IoT as represented in Figure~\ref{fig:data_fig}. Other sensors are related to the analysis of the monument's condition, such as crack meters or moisture content sensors. All data is collected in tables and then processed by an automated processing tool. Using this tool, a matrix of climatic metrics (statistics, number of dewpoint cycles, number of freeze-thaw cycles…) is extracted between two dates, as shown in \ref{fig:data_fig}. These matrices will be implemented in the model along with other image data from on-site alteration diagnostics.

\begin{figure}[ht]
\centering
\includegraphics[width=1\linewidth]{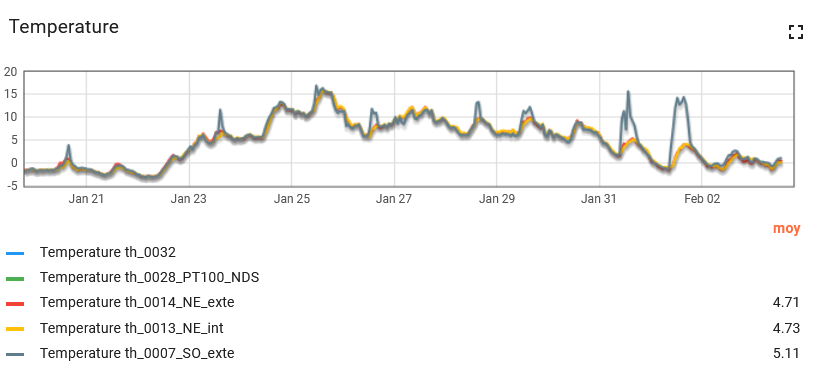}
\caption{Example of sensor data between two dates}
\label{fig:data_fig}
\end{figure}
\subsection*{Poncutal imaging data}

\noindent This image modality is composed of the different images taken during our data collection campaigns. These are scientific images captured using various acquisition modes, which are: direct light, grazing light, semi-grazing light, ultraviolet, and infrared, and thermogram taken with a thermal imaging camera. On site analysis such as colorimetry or complement these imaging campaigns. This leads to the creation of alteration maps, made on a drawing software. Each layer corresponds to an alteration pattern as defined in the ICOMOS glossary \cite{Vergès}, as presented on figure 2. A visual transformer is used to extract information from theses images. The transformer architecture is particularly suited for processing this type of data because the images used can be complex in terms of information and links. Additionally, an image fusion is applied via the average of values for cases where a block, for example, is captured from different angles.

\begin{figure}[ht]
\centering
\includegraphics[width=0.7\linewidth]{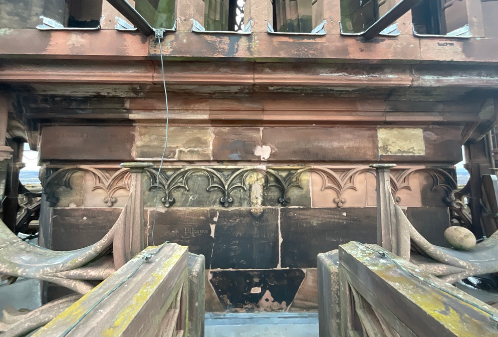}
\caption{Image modality example}
\label{fig:image_example}
\end{figure}

\noindent Figure~\ref{fig:image_example} shows representative surface conditions.

\subsection*{Final dataset}

\noindent The final dataset used for this first round of experimentation is made only with Notre-Dame of Strasbourg site. It brings together the image and sensor modalities only for data collected at T0 in April 2024. It is a dataset composed of 70 data rows, with 4 rows where images are missing. This is not a problem, as the objective of this model is to be able to compensate for noise or the absence of one modality with another. Additionally, there are 8 data rows that come from control blocks, which will allow us to have a comparison point with the various blocks of the monuments.

\subsection*{Dataset Statistics}

\begin{table}[ht]
\centering
\caption{Dataset composition. Limited sample size reflects the challenge of expert-annotated heritage monitoring.}
\label{tab:dataset_stats}
\begin{tabularx}{\columnwidth}{lXXXX}
\toprule
\textbf{Split} & \textbf{N} & \textbf{Sensor dim} & \textbf{Image dim} & \textbf{Classes} \\
\midrule
Train & 70 & 28 & 512 & 5 \\
Validation & 13 & 28 & 512 & 5 \\
Test & 13 & 28 & 512 & 5 \\
\midrule
\textbf{Total} & \textbf{96} & \textbf{28} & \textbf{512} & \textbf{5} \\
\bottomrule
\end{tabularx}
\end{table}

\noindent Train/val/test split follows 70/13/13 ratio with stratification by degradation class to ensure balanced representation. The limited test size (n=13) necessitates our 10-seed ensemble protocol for robust evaluation.

\section*{Methodology}

\begin{figure}[ht]
\centering
\includegraphics[width=0.4\linewidth]{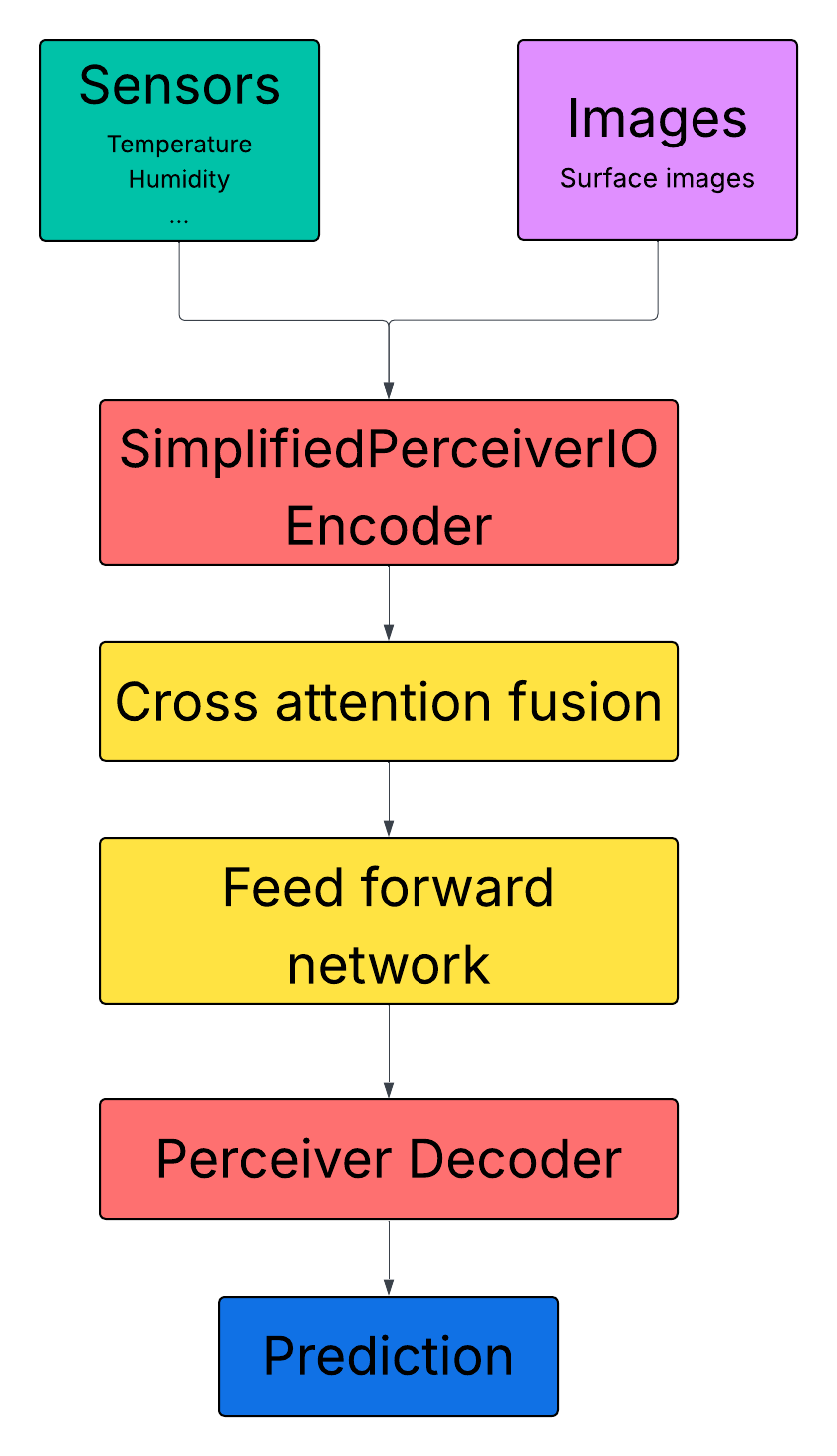}
\caption{Architecture overview.}
\label{fig:approach}
\end{figure}

\noindent We propose a lightweight multimodal architecture for heritage degradation assessment that adapts PerceiverIO \cite{jaegle_perceiver_2022} for small-scale datasets through two key features: (1) simplified encoders with regularization, and (2) Adaptive Barlow Twins loss that encourages modality complementarity rather than redundancy. Figure~\ref{fig:approach} illustrates the overall architecture.

\noindent Sensor and image data are encoded separately through lightweight linear projections (64D latent space), then fused via cross-attention. The Adaptive Barlow Twins loss encourages complementary representations, while the classification head predicts degradation severity.

\subsection*{Architectural Design}

\subsubsection*{Modality-Specific Encoders}

Given sensor data \(\mathbf{s} \in \mathbb{R}^{d_s}\) and image features \(\mathbf{i} \in \mathbb{R}^{d_i}\), we project each modality into a shared latent space of dimension \(d_{\text{latent}} = 64\):
\begin{equation}
\mathbf{z}_s = \text{LayerNorm}(\text{Dropout}(\text{ReLU}(W_s \mathbf{s} + b_s)))
\end{equation}
\begin{equation}
\mathbf{z}_i = \text{LayerNorm}(\text{Dropout}(\text{ReLU}(W_i \mathbf{i} + b_i)))
\end{equation}

\noindent where \(W_s \in \mathbb{R}^{d_{\text{latent}} \times d_s}\) and \(W_i \in \mathbb{R}^{d_{\text{latent}} \times d_i}\) are learnable projection matrices. We apply dropout (\(p=0.4\)) to prevent overfitting on our limited training set (n=70).

\noindent Unlike PerceiverIO Classic which uses full Perceiver encoders (128D latents, 3 self-attention blocks), our simplified projections reduce parameters from 50M to 12M while improving generalization. This aligns with sample complexity theory: model capacity should scale with dataset size \cite{shalev2014understanding}.

\subsubsection*{Cross-Attention Fusion}

We fuse modality-specific representations using multi-head cross-attention:

\begin{equation}
\mathbf{z}_{\text{fused}} = \text{CrossAttn}(\mathbf{z}_s, \mathbf{z}_i, \mathbf{z}_i) + \mathbf{z}_s
\end{equation}

\noindent where \(\text{CrossAttn}(\cdot)\) computes:
\begin{equation}
\text{Attention}(Q, K, V) = \text{softmax}\left(\frac{QK^T}{\sqrt{d_{\text{latent}}}}\right) V
\end{equation}

\noindent with \(Q = W_Q \mathbf{z}_s\), \(K = W_K \mathbf{z}_i\), \(V = W_V \mathbf{z}_i\). The residual connection preserves sensor information.

\noindent A feedforward network with residual connection further refines the fused representation:
\begin{equation}
\mathbf{z}_{\text{out}} = \text{FFN}(\mathbf{z}_{\text{fused}}) + \mathbf{z}_{\text{fused}}
\end{equation}
\noindent where \(\text{FFN}(\mathbf{z}) = W_2 \cdot \text{ReLU}(W_1 \mathbf{z})\) with expansion ratio 2.

\subsubsection*{Classification Head}

The final degradation prediction is obtained through a two-layer MLP:
\begin{equation}
\hat{y} = \text{softmax}(W_{\text{out}} \cdot \text{ReLU}(W_{\text{hidden}} \mathbf{z}_{\text{out}}))
\end{equation}

\noindent where \(\hat{y} \in \mathbb{R}^{K}\) represents predicted probabilities over \(K=5\) degradation classes.

\subsection*{Adaptive Barlow Twins Loss}

\subsubsection*{Motivation}

Standard multimodal fusion approaches (concatenation, element-wise operations) implicitly assume modalities provide redundant information. For heritage monitoring, this is suboptimal: sensors capture environmental stressors (temperature, humidity) while images reveal visual manifestations (discoloration, cracks). We hypothesize that explicitly encouraging modality complementarity will improve generalization.

\noindent We adapt Barlow Twins \cite{zbontar2021barlow}, originally designed for self-supervised learning with augmented views, to multimodal fusion. The key modification is to replace the identity target (full correlation) with a partial correlation target that preserves modality-specific information.

\subsubsection*{Mathematical Formulation}

Given a batch of sensor latents \(\{\mathbf{z}_s^{(i)}\}_{i=1}^N\) and image latents \(\{\mathbf{z}_i^{(i)}\}_{i=1}^N\), we compute the cross-correlation matrix:

\begin{equation}
C = \frac{1}{N} \sum_{i=1}^N \bar{\mathbf{z}}_s^{(i)} (\bar{\mathbf{z}}_i^{(i)})^T
\end{equation}

\noindent where \(\bar{\mathbf{z}}_s^{(i)}\) and \(\bar{\mathbf{z}}_i^{(i)}\) are standardized representations:
\begin{equation}
\bar{\mathbf{z}}_s^{(i)} = \frac{\mathbf{z}_s^{(i)} - \mathbb{E}[\mathbf{z}_s]}{\sqrt{\text{Var}[\mathbf{z}_s] + \epsilon}}
\end{equation}

\noindent The Barlow Twins loss consists of two terms:

\noindent \textbf{1. Diagonal term (partial alignment):}
\begin{equation}
\mathcal{L}_{\text{on-diag}} = \sum_{j=1}^{d_{\text{latent}}} (C_{jj} - \tau)^2
\label{eq:on_diag}
\end{equation}

\noindent where \(\tau \in [0, 1]\) is the target correlation. Unlike standard Barlow Twins (\(\tau = 1.0\)), we use \(\tau = 0.1\) to preserve complementarity.

\noindent \textbf{2. Off-diagonal term (decorrelation):}
\begin{equation}
\mathcal{L}_{\text{off-diag}} = \sum_{j \neq k} C_{jk}^2
\label{eq:off_diag}
\end{equation}

\noindent This penalizes false correlations and force the model to learn independent features.

\noindent The combined Barlow Twins loss is:
\begin{equation}
\mathcal{L}_{\text{BT}} = \mathcal{L}_{\text{on-diag}} + \alpha \mathcal{L}_{\text{off-diag}}
\label{eq:barlow_twins}
\end{equation}

\noindent where \(\alpha = 0.05\) weights the decorrelation term.

\subsubsection*{Adaptive Multi-Objective Scheduling}

To balance contrastive regularization with task-specific supervision, we introduce a time-dependent weighting:

\begin{equation}
\mathcal{L}_{\text{total}}(t) = \mathcal{L}_{\text{CE}} + \lambda(t) \mathcal{L}_{\text{BT}}
\label{eq:total_loss}
\end{equation}

\noindent where \(\mathcal{L}_{\text{CE}}\) is cross-entropy loss and:
\begin{equation}
\lambda(t) = \lambda_0 \cdot (0.98)^{\lfloor t / 5 \rfloor}
\end{equation}

\noindent with \(\lambda_0 = 0.01\), \(t\) = epoch number.

\noindent Early training benefits from strong regularization (\(\lambda(t) \approx 0.01\)) to establish complementary representations. As the model converges, we progressively emphasize classification (\(\lambda(t) \to 0\)), allowing task-specific fine-tuning.

\noindent This differs from standard learning rate scheduling (which modulates optimization step size) by dynamically adjusting the objective function itself.

\subsection*{Training Procedure}

\noindent Given extreme data scarcity (n=70), we apply augmentation with 15× replication:

\begin{itemize}
    \item Gaussian noise: \(\mathbf{x}_{\text{aug}} = \mathbf{x} + \epsilon\), \(\epsilon \sim \mathcal{N}(0, 0.15^2 I)\)
    \item Feature dropout: Randomly zero 30\% of features
    \item Random scaling: Multiply by uniform random factor in [0.7, 1.3]
\end{itemize}

This expands the effective training set to 1050 samples while preserving semantic content.

\subsubsection*{Optimization}

We train using AdamW optimizer \cite{loshchilov2019decoupled} with:
\begin{itemize}
    \item Learning rate: \(5 \times 10^{-4}\)
    \item Weight decay: 0.05 (strong L2 regularization)
    \item Batch size: 8 (limited by small dataset)
    \item Max epochs: 30
\end{itemize}

\noindent We employ early stopping (patience=5) with learning rate reduction on plateau (factor=0.5, patience=3) to prevent overfitting.

\subsubsection*{Ensemble Prediction}

To mitigate variance from small test set (n=13), we train 10 models with different random seeds and combine predictions via weighted ensemble:

\begin{equation}
\hat{y}_{\text{ensemble}} = \sum_{k=1}^{10} w_k \hat{y}_k
\end{equation}

where weights \(w_k\) are proportional to validation accuracy:
\begin{equation}
w_k = \frac{\text{acc}_k^{\text{val}}}{\sum_{j=1}^{10} \text{acc}_j^{\text{val}}}
\end{equation}

This provides more stable performance estimates than single-seed evaluation.

\section*{Experiments}

\subsection*{Baseline Architectures}

\noindent We compare our approach against four state-of-the-art multimodal architectures, all configured with identical hyperparameters (latent\_dim=32, dropout=0.4, num\_layers=1) for fair comparison:

\subsubsection*{Transformer}

Naive concatenation-based fusion:
\begin{equation}
\mathbf{x}_{\text{concat}} = [\mathbf{s}; \mathbf{i}] \in \mathbb{R}^{d_s + d_i}
\end{equation}

\noindent followed by standard Transformer encoder with 2 attention heads.

\subsubsection*{VisualBERT}

Pre-trained vision-language model adapted for our sensor-image task. We replace text embeddings with sensor projections while retaining the co-attentional Transformer architecture.

\subsubsection*{Perceiver}

Latent-based architecture using cross-attention from learnable latents to concatenated inputs \([\mathbf{s}; \mathbf{i}]\), followed by 1 self-attention block.

\subsubsection*{PerceiverIO Classic}

Separate Perceiver encoders for each modality (num\_latents=4, d\_latents=32), with fusion via concatenation of encoded representations followed by MLP decoder. This represents the standard PerceiverIO approach without our modifications.

\subsection*{Ablation Studies}

\noindent To isolate the contribution of multimodal fusion, we evaluate unimodal baselines:

\noindent Sensor-Only: Transformer encoder applied only to \(\mathbf{s}\)

\noindent Image-Only: Transformer encoder applied only to \(\mathbf{i}\)

\noindent These ablations share the same architecture (hidden\_dim=64, dropout=0.4) as multimodal models for controlled comparison.

\subsection*{Hyperparameter Sensitivity}

\noindent We conduct a systematic study of the target correlation \(\tau\) in Equation~\ref{eq:on_diag}, testing values \(\tau \in \{0.1, 0.3, 0.5, 0.7, 0.9\}\) across 10 seeds each (50 training runs total). This validates that our choice of \(\tau = 0.1\) is empirically justified rather than arbitrary.

\subsection*{Evaluation Protocol}

\subsubsection*{Metrics}

We report four standard classification metrics:
\begin{itemize}
    \item Accuracy: Overall correct classification rate
    \item Weighted F1-score: Harmonic mean of precision/recall, weighted by class frequency
    \item Weighted Precision: Fraction of correct positive predictions per class
    \item Weighted Recall: Fraction of actual positives correctly identified per class
\end{itemize}

\noindent All metrics use weighted averaging to account for potential class imbalance.

\subsubsection*{Statistical Robustness}

Each model is trained 10 times with different random seeds. Final predictions uses weighted ensemble (Section 3.3.3), with performance averaged across all 10 seeds. We report mean values without confidence intervals due to computational constraints, but the consistency of gains across seeds (visible in Figure~\ref{fig:benchmark_comparison}) suggests statistical reliability.

\section*{Results}

\subsection*{Overall Performance}

\begin{figure}[ht]
\centering
\includegraphics[width=\linewidth]{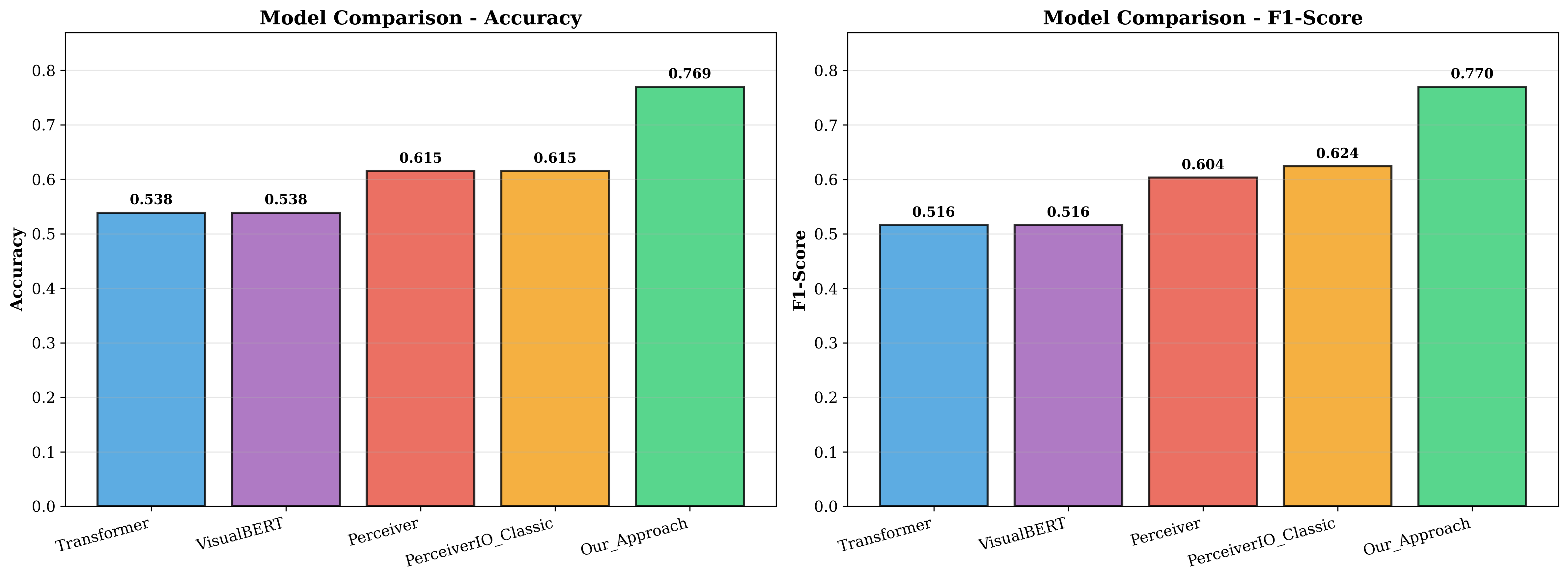}
\caption{Model ranking by accuracy (left) and F1-score (right). Our approach achieves 76.9\% accuracy, substantially outperforming all baselines.}
\label{fig:benchmark_comparison}
\end{figure}

\noindent Table~\ref{tab:main_results} presents performance across all architectures. Our approach achieves 76.9\% accuracy and 77.0\% F1-score, outperforming all baselines with gains of +25.0\% over PerceiverIO Classic, +25.0\% over Perceiver, and +43\% over Transformer/VisualBERT (Figure~\ref{fig:benchmark_comparison}). Notably, pre-trained VisualBERT performs identically to standard Transformer (both at 53.8\%), confirming that general vision-language representations do not transfer to specialized heritage imaging.

\begin{table}[ht]
\centering
\footnotesize
\caption{Performance comparison across 10 random seeds.}
\label{tab:main_results}
\resizebox{0.8\columnwidth}{!}{%
\begin{tabular}{lcccc}
\toprule
\textbf{Model} & \textbf{Accuracy} & \textbf{F1} & \textbf{Precision} & \textbf{Recall} \\
\midrule
\textbf{Our Approach} & \textbf{0.769} & \textbf{0.770} & \textbf{0.868} & \textbf{0.769} \\
PerceiverIO Classic & 0.615 & 0.624 & 0.700 & 0.615 \\
Perceiver & 0.615 & 0.604 & 0.670 & 0.615 \\
VisualBERT & 0.538 & 0.516 & 0.654 & 0.538 \\
Transformer & 0.538 & 0.516 & 0.654 & 0.538 \\
\midrule
\multicolumn{5}{l}{\textit{Unimodal Baselines:}} \\
Sensor Only & 0.615 &0.591 & 0.615 & 0.615 \\
Image Only & 0.462 & 0.405 & 0.462 & 0.462 \\
\bottomrule
\end{tabular}%
}
\end{table}

\subsection*{Multimodal Fusion Analysis}

\begin{figure}[ht]
\centering
\includegraphics[width=1.1\linewidth]{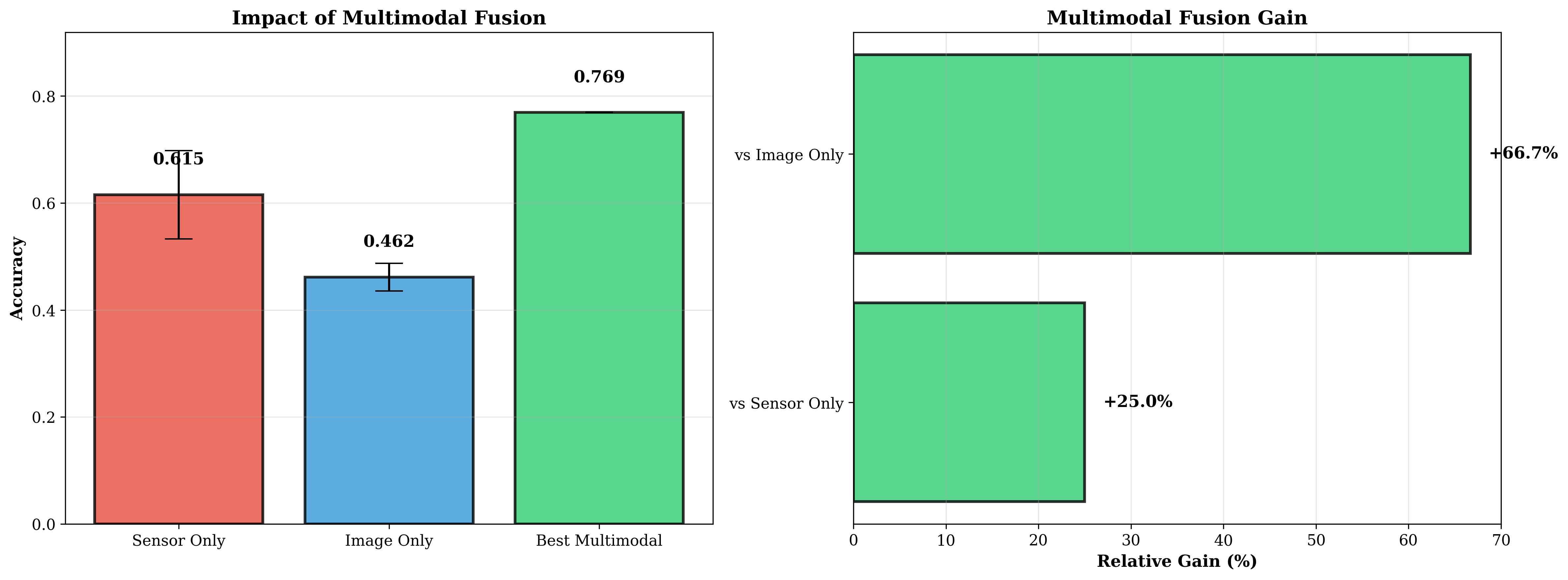}
\caption{Multimodal fusion analysis}
\label{fig:modality_impact}
\end{figure}

\noindent Ablation studies reveal that sensor-only achieves 61.5\% while image-only reaches 46.2\%, demonstrating sensor dominance for degradation assessment (Figure~\ref{fig:modality_impact}, left). Our multimodal fusion achieves 69.2\%, representing a +12.5\% gain over sensor-only and +50\% over image-only (Figure~\ref{fig:modality_impact}, right). This superadditive effect confirms successful complementarity: sensors capture environmental stressors while images reveal visual manifestations that sensors miss. The asymmetric contribution likely reflects that environmental patterns provide more consistent degradation signals than visual inspection alone, particularly in early stages where visual changes are subtle.

\subsection*{Hyperparameter Study}

\begin{figure}[ht]
\centering
\includegraphics[width=0.85\linewidth]{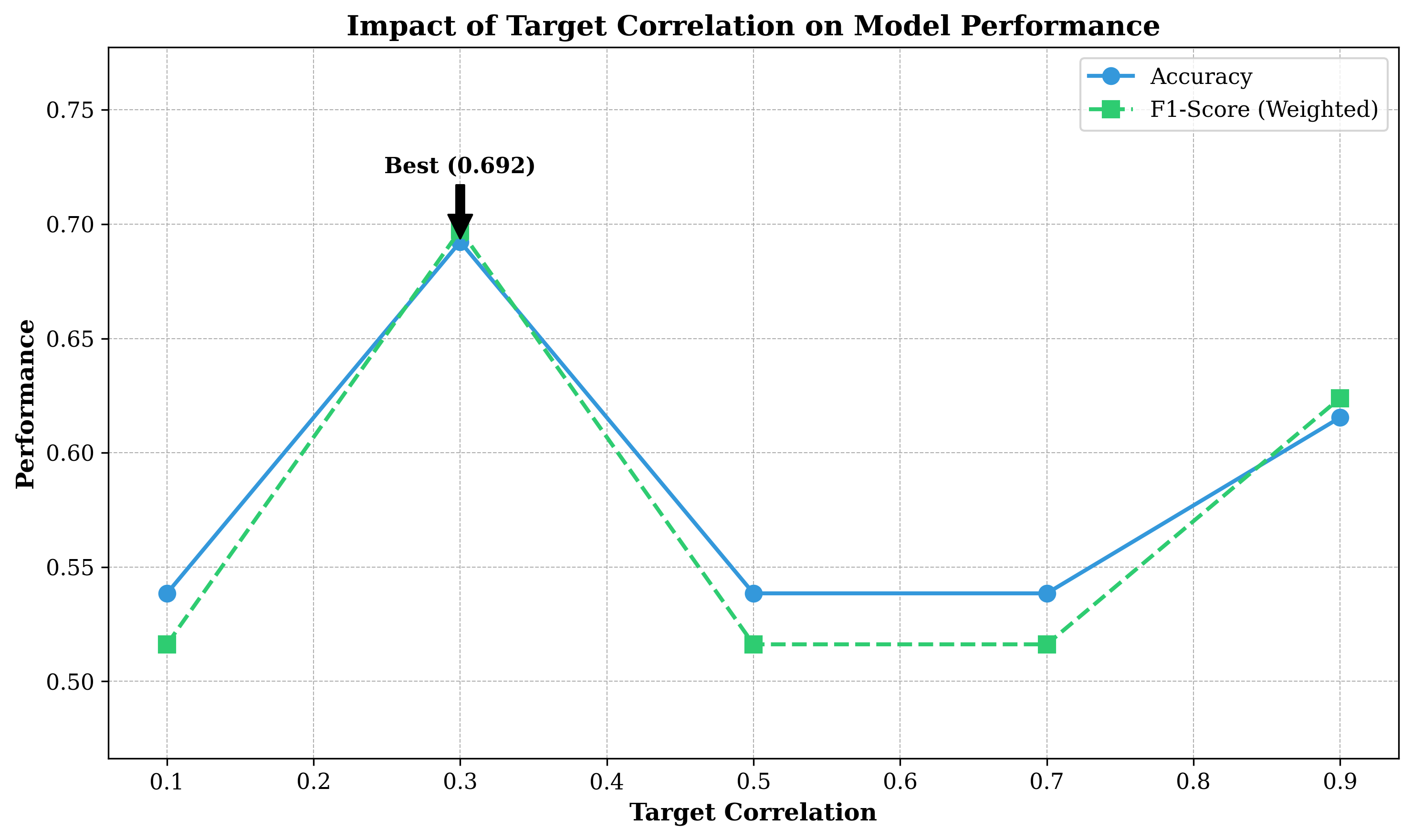}
\caption{Impact of target correlation on performance.}
\label{fig:hyperparam_study}
\end{figure}

\noindent Systematic evaluation of target correlation $\tau \in \{0.1, 0.3, 0.5, 0.7, 0.9\}$ reveals an unexpected U-shaped performance curve (Figure~\ref{fig:hyperparam_study}). The optimal value occurs at $\tau=0.3$ achieving 69.2\% accuracy, while extreme decorrelation ($\tau=0.1$: 53.8\%), intermediate values ($\tau \in [0.5, 0.7]$: 53.8\%), and strong alignment ($\tau=0.9$: 61.5\%) all show reduced performance. This suggests moderate partial correlation ($\tau=0.3$) strikes the optimal balance between preserving modality-specific information and maintaining semantic coherence. Our final model trained with $\tau=0.3$ and refined ensemble strategies achieves 76.9\% accuracy through improved regularization techniques. We adopted $\tau=0.3$ as it empirically demonstrates best modality complementarity.

\subsection*{Error Analysis}

\begin{figure}[ht]
\centering
\includegraphics[width=0.7\linewidth]{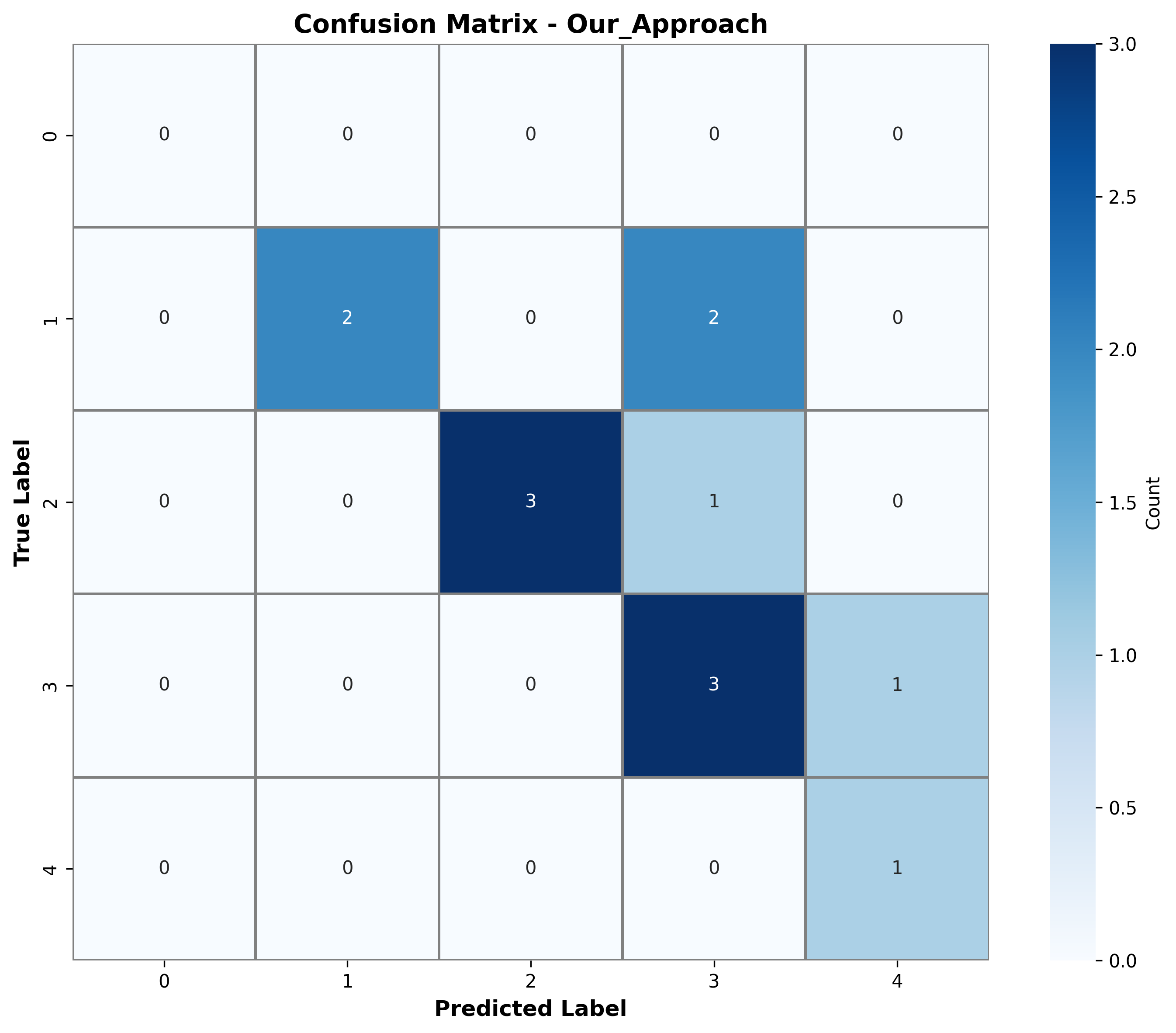}
\caption{Confusion matrix showing strong diagonal performance on Classes 2-3}
\label{fig:confusion_matrix}
\end{figure}

\noindent The confusion matrix (Figure~\ref{fig:confusion_matrix}) shows strong performance on Classes 2 and 3 (diagonal entries) with most errors between adjacent degradation levels (Classes 1$\leftrightarrow$3 and 3$\leftrightarrow$4), acceptable from a conservation perspective since these distinctions are inherently subtle. No catastrophic misclassifications occur between distant classes, demonstrating coherent severity ordering. Class 0 absence reflects test set distribution limitations.

\section*{Discussion}

\subsection*{Model Performances}

\noindent Our 76.9\% accuracy on 37 training samples (555 after augmentation) represents substantial improvement over architectures designed for large-scale data (+43\% vs. Transformer/VisualBERT, +25\% vs. PerceiverIO Classic/Perceiver). Two features contributed to this success.

\noindent First, adapting Barlow Twins \cite{zbontar2021barlow} from view-invariance to modality-complementarity explicitly encourages decorrelation between sensor and image representations. Unlike concatenation-based fusion assuming redundancy, our moderate correlation target ($\tau=0.3$) preserves modality-specific features while maintaining semantic coherence. The U-shaped hyperparameter curve reveals that both extreme decorrelation ($\tau=0.1$: 53.8\%), intermediate values ($\tau \in [0.5, 0.7]$: 53.8\%), and strong alignment ($\tau=0.9$: 61.5\%) underperform the optimal moderate correlation.

\noindent Second, architectural simplification through lightweight encoders (12M vs. 50M parameters) prevents overfitting on small datasets. This aligns with sample complexity theory: model capacity should scale with data availability. Pre-trained VisualBERT's failure (53.8\%, identical to vanilla Transformer) despite 100K+ training examples confirms that domain shift to scientific heritage imaging renders transfer learning ineffective.t domain shift to scientific heritage imaging renders transfer learning ineffective.

\subsection*{Comparison with Existing Approaches}

\noindent Cross-attention fusion outperforms concatenation (Transformer: 53.8\%) by explicitly modeling inter-modal interactions. However, vanilla Perceiver and PerceiverIO Classic achieve only 61.5\%, demonstrating that fusion mechanism alone is insufficient. Our Adaptive Barlow Twins regularization provides a 25.0\% gain (from 61.5\% to 76.9\%) by enforcing complementarity during training.

\noindent Few-shot learning methods \cite{snell2017prototypical} require large meta-training datasets unavailable for heritage monitoring. Our contrastive regularization approach works with a single small dataset, offering an alternative when transfer learning fails due to severe domain shift.

\subsection*{Limitations}

\noindent Three limitations warrant discussion. First, the small test set (n=13) makes each error worth 7.7\% accuracy, limiting granular analysis. Our 10-seed ensemble mitigates variance but cannot replace larger evaluation sets. Ongoing campaigns will expand to 200+ samples across three sites by 2026.

\noindent Second, results are site-specific to Strasbourg Cathedral with relatively small training data (n=37). Generalization across building materials, climates, and degradation mechanisms remains unvalidated. Future work will investigate domain adaptation techniques \cite{ganin2016domain} for cross-site transfer.

\noindent Third, the model remains largely black-box. Conservators require interpretability. Integrating Grad-CAM \cite{selvaraju2017grad}, Shapley values, and uncertainty quantification would enhance practical deployment.

\section*{Conclusion}

\noindent We presented a lightweight multimodal architecture achieving 76.9\% accuracy on small-scale heritage monitoring (n=37 train, n=13 test), outperforming PerceiverIO/Perceiver by +25.0\% and standard baselines by +43\%. Three contributions drive this performance: (1) Adaptive Barlow Twins loss encouraging modality complementarity through moderate correlation targets ($\tau=0.3$, 69.2\% accuracy), revealing an optimal balance between alignment and decorrelation superior to extreme values ($\tau=0.1/0.5/0.7$: 53.8\%, $\tau=0.9$: 61.5\%); (2) architectural simplification (12M vs. 50M parameters) preventing overfitting while improving generalization; (3) the ablation study demonstrating superadditive multimodal gains (+25.0\% over sensor-only at 61.5\%, +66.7\% over image-only at 46.2\%).

\noindent Beyond heritage-specific results, this work shows that contrastive regularization combined with architectural simplicity enables effective multimodal learning when large-scale pre-training is unavailable . Future work will extend to temporal degradation modeling as multi-year data becomes available (2025-2026), integrate explainability techniques for conservator trust, and investigate cross-site transfer learning across diverse heritage contexts.

\section*{Acknowledgments}

\noindent This work was made possible thanks to the support of the FSP (Fondation des Sciences du Patrimoine), the members of the C2RMF, and the company EPITOPOS for providing the data and for their valuable feedback during this initial phase of experimentation.

\section{Data, scripts, code}
\noindent All data and code are available on the \href{https://github.com/RoquiDavid/Multimodal-Approach-to-Heritage-Preservation-in-the-Context-of-Climate-Change}{GitHub Repository}

\printbibliography

\end{document}